\documentclass[twocolumn]{svjour3}
\smartqed
\usepackage[T1]{fontenc}
\usepackage{graphicx}
\usepackage{subfigure}
\usepackage{cite}
\usepackage{tipa}
\usepackage{url}
\usepackage{float}
\usepackage{amsmath,amssymb,amsfonts}
\usepackage{algorithmic}
\usepackage{graphicx}
\usepackage{textcomp}
\usepackage{adjustbox}
\usepackage[graphicx]{realboxes}
\usepackage{rotating}
\usepackage[misc]{ifsym}
\usepackage{setspace}
\usepackage{algorithm}
\usepackage{algorithmic}
\usepackage{xcolor}

\begin{document}

\title{Self-Supervised Learning for Gastritis Detection with Gastric X-ray Images}

\author{Guang Li $^{1}$ \and Ren Togo $^{2}$ \and Takahiro Ogawa $^{2}$ \and Miki Haseyama $^{2}$
}

\institute{Guang Li \\
           guang@lmd.ist.hokudai.ac.jp \\ \\
           Ren Togo \\
           togo@lmd.ist.hokudai.ac.jp \\ \\
           Takahiro Ogawa \\
           ogawa@lmd.ist.hokudai.ac.jp \\ \\
           Miki Haseyama \\
           mhaseyama@lmd.ist.hokudai.ac.jp $\textrm{\Letter}$ \\ \\
           $^{1}$ 
              Graduate School of Information Science and Technology, Hokkaido University, Sapporo, Japan \\ \\
           $^{2}$
              Faculty of Information Science and Technology, Hokkaido University, Sapporo, Japan 
}

\date{Received: date / Accepted: date}

\maketitle
\begin{abstract}
{\it Purpose:~} Manual annotation of gastric X-ray images by doctors for gastritis detection is time-consuming and expensive. To solve this, a self-supervised learning method is developed in this study. The effectiveness of the proposed self-supervised learning method in gastritis detection is verified using a few annotated gastric X-ray images.

{\it Methods:~} In this study, we develop a novel method that can perform explicit self-supervised learning and learn discriminative representations from gastric X-ray images. Models trained based on the proposed method were fine-tuned on datasets comprising a few annotated gastric X-ray images. Five self-supervised learning methods, $i.e.$, SimSiam, BYOL, PIRL-jigsaw, PIRL-rotation, and SimCLR, were compared with the proposed method. Furthermore, three previous methods, one pretrained on ImageNet, one trained from scratch, and one semi-supervised learning method, were compared with the proposed method.

{\it Results:~} The proposed method$'$s harmonic mean score of sensitivity and specificity after fine-tuning with the annotated data of 10, 20, 30, and 40 patients were 0.875, 0.911, 0.915, and 0.931, respectively. The proposed method outperformed all comparative methods, including the five self-supervised learning and three previous methods. Experimental results showed the effectiveness of the proposed method in gastritis detection using a few annotated gastric X-ray images.

{\it Conclusions:~} This paper proposes a novel self-supervised learning method based on a \textcolor{black}{teacher-student} architecture for gastritis detection using gastric X-ray images. The proposed method can perform explicit self-supervised learning and learn discriminative representations from gastric X-ray images. The proposed method exhibits potential clinical use in gastritis detection using a few annotated gastric X-ray images.
\end{abstract}
\section{Introduction}
Gastric cancer is one of the most severe malignant tumors worldwide, with the fifth highest incidence and third highest mortality rates~\cite{bray2018global}. 
The prevalence of gastric cancer is declining globally, nonetheless varying across regions~\cite{maktabi2019tissue}. 
\textcolor{black}{
The percentage of people who develop gastric cancer is higher in Eastern Asia than in other regions.
}
Among the predisposing factors of gastric cancer, chronic gastritis is considered the leading risk factor~\cite{wroblewski2010helicobacter}.
Therefore, we focus on chronic gastritis in this study.
\par
Chronic gastritis is considered the first step of gastric mucosal changes, leading to gastric cancer.
Therefore, gastritis screening has been used for identifying patients with the risk of gastric cancer~\cite{lee2016association}.
Different methods to evaluate gastritis exist, for example, double-contrast upper gastrointestinal barium X-ray radiography, upper gastrointestinal endoscopy, and serum anti-$H.\,pylori$ IgG titer~\cite{togo2019detection}.
Among such methods, the screening of gastric X-ray images is still the most simple and widely used method, and thus, it is suitable for mass screening in East Asia.
However, diagnosis requires technical knowledge and time for screening.
Computer-aided diagnosis (CAD) systems that automatically analyze gastric X-ray images and provide supporting information for physicians can overcome these problems.
\par
With the development of supervised learning based on deep convolutional neural networks (DCNNs)~\cite{lecun2015deep}, significant progress has been achieved in medical image analysis~\cite{shen2017deep}.
Gastritis detection from gastric X-ray images with high reliability using DCNN-based CAD systems has been proven in our previous studies~\cite{togo2019detection}.
Similar to most medical image analysis studies, our previous methods are based on supervised learning, and therefore, their performance depends on a large number of manually annotated gastric X-ray images~\cite{kanai2019gastritis}.
However, generating the annotations of complex gastric X-ray images typically requires expert knowledge.
Thus, it is expensive and time-consuming~\cite{zhou2021models}. 
Consequently, the scarcity of expert-level annotations is one of the main limitations impacting real-world applications in gastritis detection.
\par
In recent years, self-supervised learning has been extensively researched~\cite{liu2021self}.
Unlike supervised learning that requires manually annotated labels, self-supervised learning benefits from image characteristics ($e.g.$, color and texture) without the manual annotation of labels~\cite{jing2020self}.
Subsequently, a model trained with self-supervised learning can be used to fine-tune the gastric X-ray image dataset.
As fine-tuning after self-supervised learning requires only a small portion of annotated data, a feasible solution is provided for insufficient data annotations in gastritis detection.
For example, a previous study~\cite{gidaris2018unsupervised} predicted the rotation degrees of images, and the study~\cite{noroozi2016unsupervised} learned discriminative representations by playing a jigsaw game on images.
Because these methods are designed for ordinary images, they cannot learn sufficient discriminative representations from gastric X-ray images. 
\par
Herein, we propose a novel self-supervised learning method based on a \textcolor{black}{teacher-student} architecture for gastritis detection using gastric X-ray images.
\textcolor{black}{
We present cross-view and cross-model losses that enable explicit self-supervised learning and learn discriminative representations from gastric X-ray images.
}
Experimental results show that the proposed method achieves a high detection performance for gastritis detection using only a few annotations.
To the best of our knowledge, this is the first study to prove the effectiveness of self-supervised learning in the field of gastritis detection using gastric X-ray images.
\par
Our contributions are summarized as follows:
\begin{itemize}
    \item We propose a novel self-supervised learning method for learning discriminative representations from gastric X-ray images.
    \item We realize high detection performance on a complex gastric X-ray image dataset with only a few annotations.
\end{itemize}
\section{Methods}
\subsection{Gastric X-ray image preprocessing}
\label{}
\begin{figure*}[t]
        \centering
        \includegraphics[width=16cm]{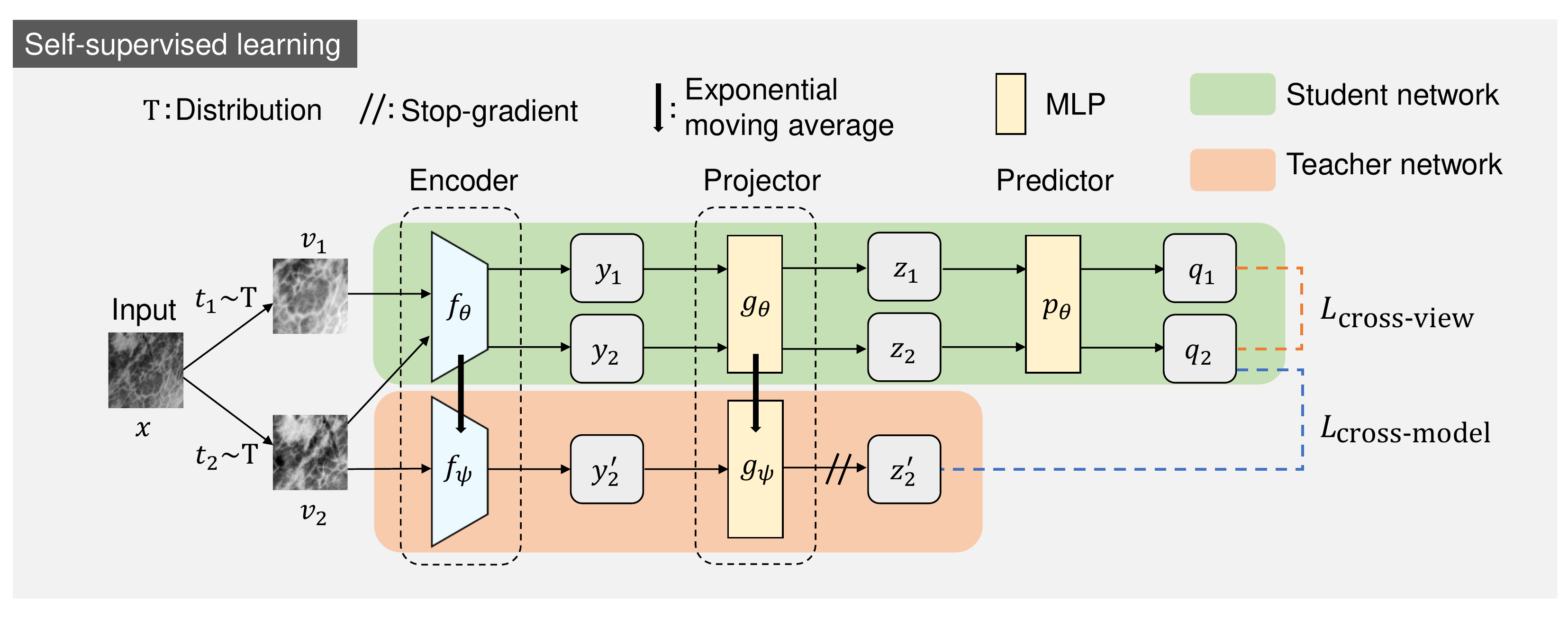}
        \caption{Overview of the proposed method. Our method minimizes (1) a cross-view loss between the gastric features of two views from the student network and (2) a cross-model loss between the gastric features of the same view from the \textcolor{black}{teacher-student} networks. $\mathrm{MLP}$ represents multilayer perceptron.}
        \label{fig1}
\end{figure*}
\begin{figure}[t]
        \centering
        \subfigure[]{
        \centering
        \includegraphics[width=3.5cm]{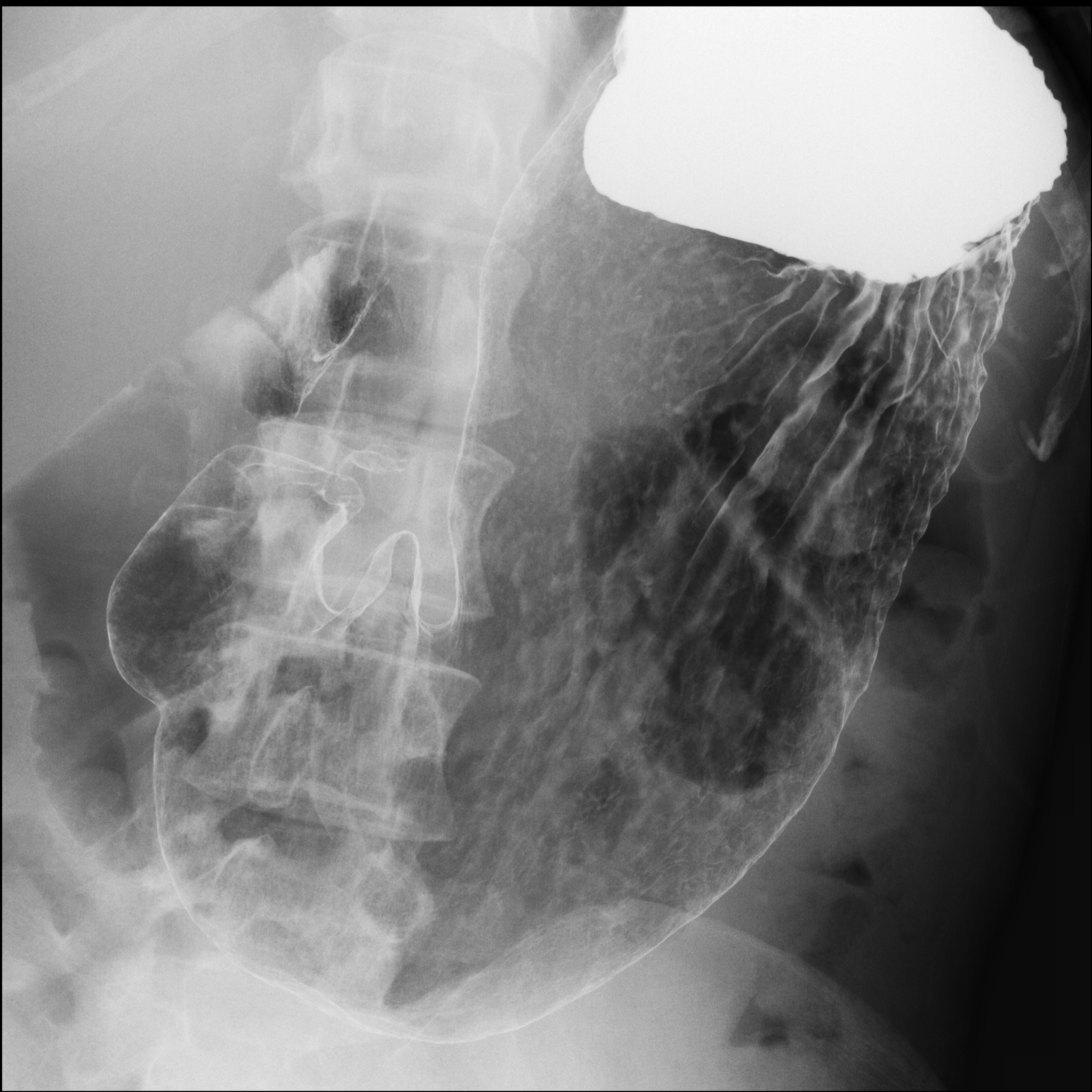}
        }
        \subfigure[]{
        \centering
        \includegraphics[width=3.5cm]{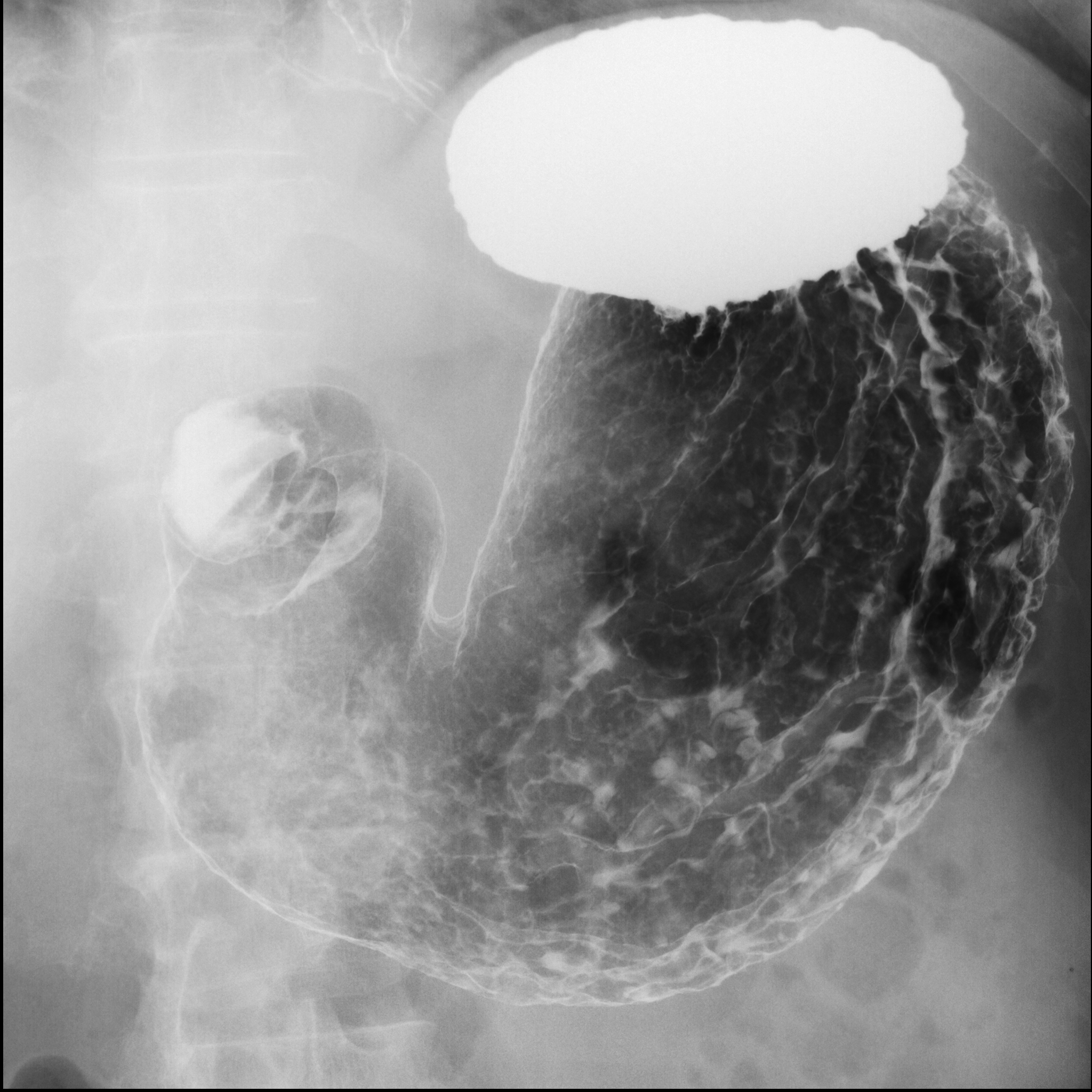}
        }
        \caption{Examples of patient-level gastric X-ray images: (a) negative image and (b) positive image.}
        \label{fig2}
\end{figure}
\begin{figure}[t]
        \centering
        \subfigure[]{
        \centering
        \includegraphics[width=2cm]{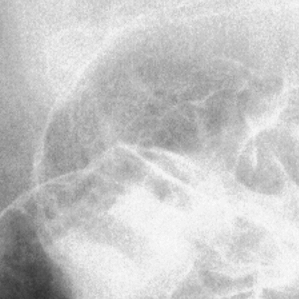}
        \includegraphics[width=2cm]{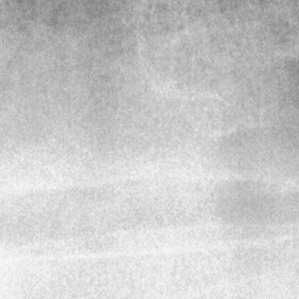}
        \includegraphics[width=2cm]{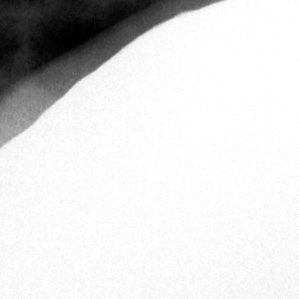}
        \includegraphics[width=2cm]{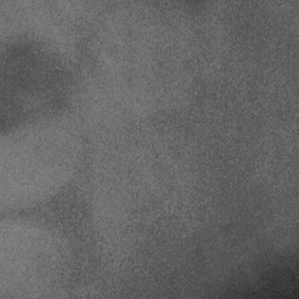}
        }
        \subfigure[]{
        \includegraphics[width=2cm]{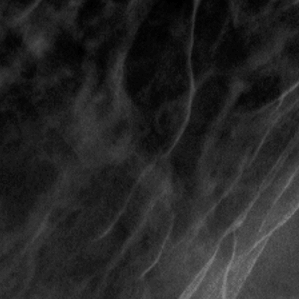}
        \includegraphics[width=2cm]{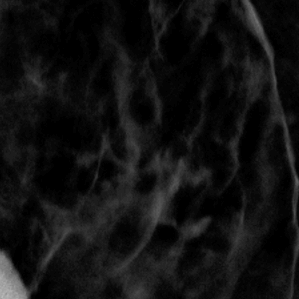}
        \includegraphics[width=2cm]{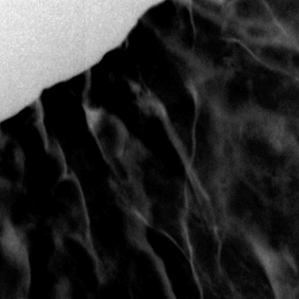}
        \includegraphics[width=2cm]{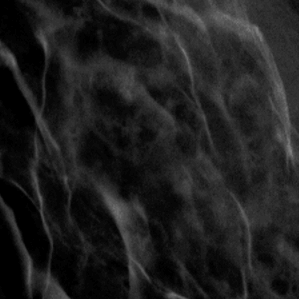}
        }
        \subfigure[]{
        \centering
        \includegraphics[width=2cm]{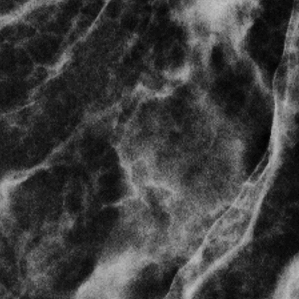}
        \includegraphics[width=2cm]{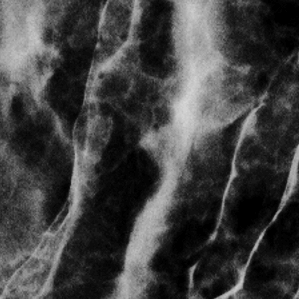}
        \includegraphics[width=2cm]{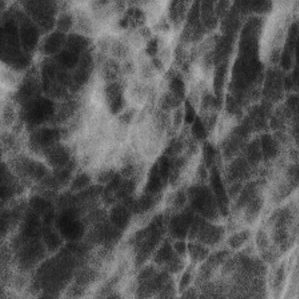}
        \includegraphics[width=2cm]{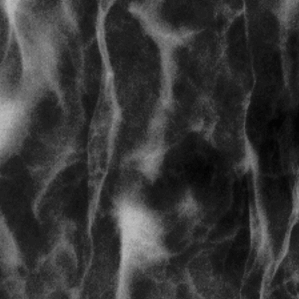}
        }
        \caption{Examples of patches: (a) outside patches in $\mathcal{O}$, (b) negative patches in $\mathcal{N}$, and (c) positive patches in $\mathcal{P}$.}
        \label{fig3}
\end{figure}
The resolution of the patient-level gastric X-ray images in our dataset is 2,048 $\times$ 2,048 pixels.
Further, the dataset contains only a few images.
To fully use the semantic information of the gastric X-ray images, we divide each patient-level image into patches and manually annotate them with the following three labels: 
\begin{itemize}
    \item $\mathcal{O}$: patches outside the stomach (outside patches), 
    \item $\mathcal{N}$: patches extracted from negative (non-gastritis) X-ray images inside the stomach (negative patches),
    \item $\mathcal{P}$: patches extracted from positive (gastritis) X-ray images inside the stomach (positive patches). 
\end{itemize}
Figure~\ref{fig2} shows examples of patient-level gastric X-ray images.
As shown in the figure, the stomach without gastritis (Figure~ \ref{fig2}-(a)) includes straight folds and uniform mucosal surface patterns. 
Conversely, the stomach with gastritis (Figure~\ref{fig2}-(b)) has non-straight folds and coarse mucosal surface patterns.
Figure~\ref{fig3} illustrates the examples of divided gastric X-ray image patches.
Figure~\ref{fig3}-(a) illustrates outside patches in $\mathcal{O}$.
Figure~\ref{fig3}-(b) displays negative patches in $\mathcal{N}$.
Figure~\ref{fig3}-(c) shows positive patches in $\mathcal{P}$.
\subsection{Self-supervised learning}
An overview of the proposed method is shown in Fig.~\ref{fig1}.
The \textcolor{black}{teacher-student} architecture comprises two networks with the same structure, where the weights of the teacher network are an exponential moving average of the weights of the student network~\cite{antti2017mean}.
Encoder $f_{\theta}$, projector $p_{\theta}$, and predictor $g_{\theta}$ belong to the student network.
Encoder $f_{\psi}$ and projector $g_{\psi}$ belong to the teacher network.
\textcolor{black}{
The teacher-student architecture is designed for learning discriminative representations ($e.g.$, gastric
folds or mucosal surfaces) from gastric patches.
}
\par
Given an input gastric X-ray patch $x$, two transformations $t_{1}$ and $t_{2}$ are randomly sampled from the distribution $\mathrm{T}$ to generate two views $v_{1} = t_{1}(x)$ and $v_{2} = t_{2}(x)$.
\textcolor{black}{
These transformations are combined with standard data augmentation methods such as cropping, resizing, flipping, and Gaussian blurring, which can increase the diversity of gastric features.
}
The view $v_{1}$ is transmitted into the encoder $f_{\theta}$ and transformed by the projector $g_{\theta}$ in the student network.
Accordingly, the view $v_{2}$ is transmitted into the encoder $f_{\psi}$ and transformed by the projector $g_{\psi}$ in the teacher network, where $z'_{2}$ is the obtained gastric feature.
\textcolor{black}{
One copy of $v_{2}$ is transmitted into the student network for calculating the final loss.
}
Further, predictor $p_{\theta}$ is used to transform two views to gastric features $q_{1}$ and $q_{2}$ in the student network.
The predictor $p_{\theta}$ and projectors $g_{\theta}$, $g_{\psi}$ are MLP with the same structure~\cite{chen2020simple}.
\textcolor{black}{
Since the computed gastric features are high-dimensional, MLP aims to reduce the dimensionality of the gastric features for fast learning.
}
\par
Finally, self-supervised learning is performed by reducing the distance between the gastric features from the student network and reducing the distance between the gastric features from the \textcolor{black}{teacher-student} networks.
\\
$\mathbf{Cross}$-$\mathbf{view}\,\,\mathbf{loss.}$
The cross-view loss defined by the following equation compares the representations of gastric features from the student network, which penalizes gastric features for different views from the same network:
\begin{equation}
\label{equ1}
\begin{split}
L_{\mathrm{cross\mathchar`-view}} 
& = || \hat{q}_{1} - \hat{q}_{2} ||_{2}^{2} 
\\ & = 2 - 2 \cdot \frac{\left \langle q_{1},q_{2} \right \rangle}{ || q_{1} ||_{2} \cdot || q_{2} ||_{2}},
\end{split}
\end{equation}
where $\hat{q}_{1} = q_{1}/ || q_{1} ||_{2}$ and $\hat{q}_{2} = q_{2}/ || q_{2} ||_{2}$ represent the normalized gastric features of $v_{1}$ and $v_{2}$ from the student network, respectively.
\\
$\mathbf{Cross}$-$\mathbf{model}\,\,\mathbf{loss.}$
The cross-model loss defined by the following equation compares the representations of the same view from the \textcolor{black}{teacher-student} networks, which penalizes gastric features for the same view from different networks:
\begin{equation}
\label{equ2}
\begin{split}
L_{\mathrm{cross\mathchar`-model}} 
& = || \hat{q}_{2} - \hat{z}'_{2} ||_{2}^{2} 
\\ & = 2 - 2 \cdot \frac{\left \langle q_{2},z'_{2} \right \rangle}{ || q_{2} ||_{2} \cdot || z'_{2} ||_{2}},
\end{split}
\end{equation}
where $\hat{z}'_{2} = z'_{2}/ || z'_{2} ||_{2}$ represents the normalized gastric features of $v_{2}$ from the teacher network.
The predictor is only used in the student network to render the \textcolor{black}{teacher-student} architecture asymmetric, which can prevent learning from collapsing~\cite{grill2020bootstrap}.
\textcolor{black}{
The consistency among views from the teacher-student networks enables learning discriminative representations ($e.g.$, gastric folds or mucosal surfaces) from the gastric patches.
}
The weights of the student network ($\theta$) are updated by minimizing the total loss.
The total loss $L_{\theta,\psi}$ and optimizing process are defined as follows:
\begin{equation}
\label{equ3}
L_{\theta,\psi} = L_{\mathrm{cross\mathchar`-view}} + L_{\mathrm{cross\mathchar`-model}},
\end{equation}
\begin{equation}
\label{equ4}
\theta \leftarrow \mathrm{Opt}(\theta, \nabla_{\theta}L_{\theta,\psi}, \alpha),
\end{equation}
$\mathrm{Opt}$ and $\alpha$ represent an optimizer and the learning rate, respectively.
The weight-updating process of the teacher network ($\psi$) is defined as follows:
\begin{equation}
\label{equ5}
\psi \leftarrow \tau\psi + (1-\tau)\theta,
\end{equation}
where $\tau$ represents the degree of moving average.
The teacher network is not updated using backpropagation, because the stop-gradient operation prevents the collapse of self-supervised learning~\cite{chen2021exploring}.
Implementing self-supervised learning, the encoder of the student network ($f_{\theta}$) can learn sufficient discriminative representations from the gastric patches and can be used for fine-tuning and gastritis detection.
\subsection{Fine-tuning and gastritis detection}
\textcolor{black}{
After network training using self-supervised learning, we fine-tune the encoder of the student network ($f_{\theta}$) with only a few annotated images, which can reduce the labor of doctors labeling.
}
In the test phase, we also divide the patient-level gastric X-ray images into patches, as described in Section 2.1.
Thereafter, we load the divided patches into the fine-tuned DCNN model and predict the labels of these patches.
Subsequently, we calculate the number of patches estimated as $\tilde{\mathcal{N}}$ and $\tilde{\mathcal{P}}$.
We do not count the number of patches estimated as $\tilde{\mathcal{O}}$ because these patches outside the stomach are not related to the final gastritis detection.
Finally, we estimate the label of a patient-level gastric X-ray image as follows:
\begin{equation}
y^\mathrm{test} = 
\left\{\begin{matrix}
 \mathrm{1} & \mathrm{if} \, \frac{\tilde{\mathcal{P}}}{\tilde{\mathcal{N}} \,+\, \tilde{\mathcal{P}}} \geq \sigma \hfill \\
 \mathrm{0} & \mathrm{otherwise} \hfill
\end{matrix}\right.
,
\end{equation}
where $\tilde{\mathcal{N}}$ and $\tilde{\mathcal{P}}$ represent the number of estimated negative and positive patches, respectively.
$\sigma$ represents a threshold that can be adjusted according to different experimental conditions.
If $y^\mathrm{test} = 1$, the estimated label of a patient-level gastric X-ray image is positive, and the estimated label is negative if $y^\mathrm{test} = 0$.
\begin{figure}[t!]
    \begin{center}
        \includegraphics[width=8cm]{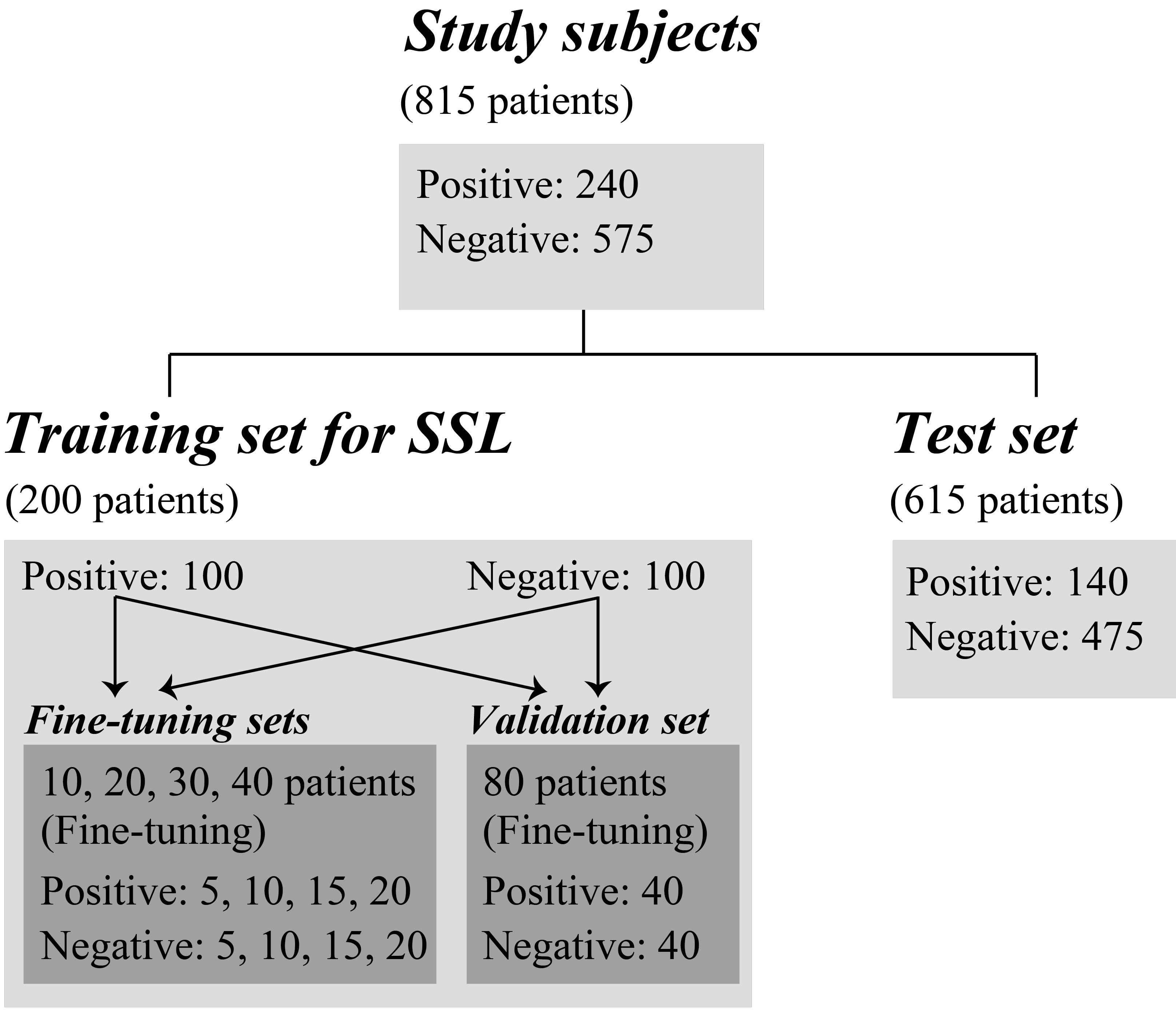}
        \caption{Details of the partitioned datasets used in the present study. SSL represents the self-supervised learning process.}
        \label{fig4}
    \end{center}
\end{figure}
\section{Experiments}
\begin{table*}[t!]
    \small
    \centering
    \caption{Comparison with different self-supervised learning methods.}
    \label{tab2}
    \begin{tabular}{c|ccc|ccc|ccc|ccc}
    \hline
    & & 10 patients & & & 20 patients & & & 30 patients & & & 40 patients &\\\hline
    Method & Sen & Spe & HM & Sen & Spe & HM & Sen & Spe & HM & Sen & Spe & HM \\\hline\hline
    PM 
    & 0.957 & 0.806 & 0.875 
    & 0.964 & 0.863 & 0.911 
    & 0.936 & 0.895 & 0.915   
    & 0.964 & 0.901 & 0.931 \\
    CM1 
    & 0.950 & 0.739 & 0.831 
    & 0.964 & 0.771 & 0.857 
    & 0.921 & 0.863 & 0.891   
    & 0.907 & 0.928 & 0.918 \\
    CM2 
    & 0.964 & 0.758 & 0.849
    & 0.907 & 0.920 & 0.913
    & 0.807 & 0.956 & 0.875   
    & 0.964 & 0.861 & 0.910 \\
    CM3
    & 0.721 & 0.806 & 0.761
    & 0.943 & 0.507 & 0.659
    & 0.879 & 0.743 & 0.805
    & 0.864 & 0.737 & 0.795 \\
    CM4
    & 0.607 & 0.735 & 0.665
    & 0.929 & 0.316 & 0.472
    & 0.879 & 0.440 & 0.586
    & 0.886 & 0.632 & 0.738 \\
    CM5 
    & 0.707 & 0.274 & 0.395  
    & 0.521 & 0.617 & 0.565 
    & 0.407 & 0.762 & 0.531   
    & 0.479 & 0.861 & 0.615 \\
    \hline
    \end{tabular}
\end{table*}
\subsection{Dataset}
\begin{table*}[t!]
    \centering
    \caption{Comparison with our previous methods. ``-" represents there are no results reported in~\cite{li2020chronic}.}
    \label{tab3}
    \begin{tabular}{c|c|c|c|c||c}
    \hline
    Method & 10 patients & 20 patients & 30 patients & 40 patients & 200 patients \\\hline\hline
    PM 
    & 0.875 
    & 0.911 
    & 0.915   
    & 0.931
    & 0.954 \\
    CM6
    & 0.860
    & 0.870
    & -   
    & -
    & 0.922 \\
    CM7
    & 0.382 
    & 0.759 
    & 0.435   
    & 0.870
    & 0.954 \\
    CM8 
    & 0.348
    & 0.477
    & 0.563
    & 0.644
    & 0.876 \\
    \hline
    \end{tabular}
\end{table*}
\begin{table}[t!]
    \small
	\begin{center}
		\caption{Hyperparameters of the proposed method.}
		\label{tab1}
		\begin{tabular}[t]{lc}
			\hline
			 Parameter  & Value \\
			\hline\hline
             Epoch         & 80 \\
             Batch size    & 256 \\
		     Learning rate ($\alpha$) & 0.03 \\
		     momentum      & 0.9 \\
		     weight decay  & 0.0004 \\
			 moving average ($\tau$) & 0.996 \\
			 mlp hidden size & 4096 \\
			 projection size & 256 \\
			 View size      & 128 \\
			 Threshold ($\sigma$) & 0.5 \\
			\hline
		\end{tabular}
	\end{center}
\end{table}
In this study, the medical data were provided by The University of Tokyo Hospital in Japan~\cite{togo2019detection}.
The medical data include gastric X-ray images of 815 patients subjects (240 positive and 575 negative).
The resolution of these gastric X-ray images is 2,048 $\times$ 2,048 pixels.
Each image has a ground truth label (positive/negative), determined based on the diagnostic results of an X-ray inspection and endoscopic examination.  
We used 200 patients$'$ images (100 positive and 100 negative) as the training set and the rest as the test set.
As shown in Section 2.1, we first divided the gastric X-ray images into patches.
The patch size and sliding interval were set to 299 and 50 pixels, respectively, as described in our previous study~\cite{kanai2019gastritis}.
The patches in the training data were annotated as $\mathcal{O}$, $\mathcal{N}$, and $\mathcal{P}$ by a radiological technologist.
Note that if the area inside the stomach in the patch was less than 1$\%$, the patch was annotated as $\mathcal{O}$.
\textcolor{black}{
If the area inside the stomach in the patch was greater than 85$\%$, the patch was annotated as $\mathcal{N}$ or $\mathcal{P}$.
}
We discarded the rest of the patches in the training data.
The number of obtained $\mathcal{O}$, $\mathcal{N}$, and $\mathcal{P}$ patches were 48,385, 42,785, and 45,127, respectively.
\par
The obtained 200 patients$'$ patches were used as the training set for the self-supervised learning process, and their label information was not used.
\textcolor{black}{
We split the 200 patients$'$ patches into 120 and 80 patients$'$ patches (half positive and half negative) as the training and validation sets, respectively, for the fine-tuning process.
}
Thereafter, we randomly selected 10, 20, 30, and 40 patients$'$ patches (half positive and half negative) from the training set as the fine-tuning sets of the fine-tuning process. 
Details of the partitioned data used in this study are shown in Figure~\ref{fig4}.
\subsection{Implementation}
For distribution $\mathrm{T}$, we used sequential data augmentation methods for random view generation, including cropping, resizing, flipping, and Gaussian blurring~\cite{li2022self, li2022tribyol}.
ResNet50~\cite{he2016deep} was used as the encoder.
The optimizer used in the present method was the SGD optimizer, whose learning rate $\alpha$, momentum, and weight decay were 0.03, 0.9, and 0.0004, respectively.
\textcolor{black}{
We performed 80 epochs in the self-supervised learning process.
}
Hyperparameters of the proposed method are shown in Table~\ref{tab1}.
\par
In the fine-tuning process, we used the weights of the trained encoder in the student network ($f_{\theta}$) as initial weights.
We assume that the effect of self-supervised learning is positively correlated with the final gastritis detection performance.
To verify the effectiveness of the proposed method (PM), we compared it with the following methods.
CMs 1--5 are several self-supervised learning methods.
CMs 6--8 are our previous methods.
\\
$\mathbf{CM1}$. We used the weights of the trained encoder of SimSiam~\cite{chen2021exploring} as initial weights for the fine-tuning process.
\\
$\mathbf{CM2}$. We used the weights of the trained encoder of BYOL~\cite{grill2020bootstrap} as initial weights for the fine-tuning process.
\\
$\mathbf{CM3}$. We used the weights of the trained encoder of PIRL~\cite{misra2020self} based on the jigsaw pretext task as initial weights for the fine-tuning process.
\\
$\mathbf{CM4}$. We used the weights of the trained encoder of PIRL~\cite{misra2020self} based on the rotation prediction pretext task as initial weights for the fine-tuning process.
\\
$\mathbf{CM5}$. We used the weights of the trained encoder of SimCLR~\cite{chen2020simple} as initial weights for the fine-tuning process.
\\
$\mathbf{CM6}$. A semi-supervised learning method based on tri-training~\cite{li2020chronic}.
\\
$\mathbf{CM7}$. A supervised learning method based on transfer learning~\cite{kanai2019gastritis}.
\\
$\mathbf{CM8}$. We used random weights (trained from scratch) as initial weights for the fine-tuning process.
\par
We performed self-supervised learning and fine-tuned the trained encoder on all four fine-tuning sets.
In all the experiments, we selected the model that exhibited the highest accuracy on the validation set and tested gastritis detection performance on the test set of 615 patients$'$ data.
Note that the settings of SimSiam and BYOL used in our experiments were strictly the same as those in the PM.
The settings of PIRL and SimCLR used in our experiments were based on the suggestions in the original papers~\cite{misra2020self, chen2020simple}.
\textcolor{black}{
We used the same settings in the fine-tuning process across all the experiments.
}
\par
In the test phase, we experimentally set the threshold $\sigma$ to 0.5 for realizing a high gastric detection performance.
We used sensitivity (Sen), specificity (Spe), and harmonic mean (HM) as the evaluation metrics.
\begin{equation}
\mathrm{Sen} = \frac{\mathrm{TP}}{\mathrm{TP + FN}},
\end{equation}
\begin{equation}
\mathrm{Spe} = \frac{\mathrm{TN}}{\mathrm{TN + FP}},
\end{equation}
\begin{equation}
\mathrm{HM} = \frac{\mathrm{2 \times Sen \times Spe}}{\mathrm{Sen + Spe}},
\end{equation}
where TP, TN, FP, and FN represent the number of true positive, true negative, false positive, and false negative, respectively. 
As Sen and Spe exhibit a tradeoff relationship, we consider HM as the final evaluation metric. 
\subsection{Results}
The experimental results are shown in Table~\ref{tab2} and~\ref{tab3}.
Table~\ref{tab2} shows the patient-level gastritis detection results after fine-tuning with the annotated data of 10, 20, 30, and 40 patients in the experiments.
According to Table~\ref{tab2}, the PM outperforms other comparative methods in terms of gastritis detection.
The PM$'$s HM score of sensitivity and specificity after fine-tuning with the annotated data of 10, 20, 30, and 40 patients are 0.875, 0.911, 0.915, and 0.931, respectively.
The average HM scores of the PM are greater than those of the other self-supervised learning methods (CMs1--5) with 0.034, 0.021, 0.153, 0.293, and 0.351 on four randomly selected training sets, respectively.
Experimental results showed that our method achieves a high gastritis detection performance with only a few annotations, which can substantially reduce the number of manually annotated labels required. 
\par
Table~\ref{tab3} shows the patient-level gastritis detection results after fine-tuning with different numbers of the annotated data patients for comparison with our previous methods.
For semi-supervised learning based on tri-training, we directly use the results reported in~\cite{li2020chronic} for a reasonable comparison.
As per Table~\ref{tab3}, our method not only drastically outperforms previous methods with a small amount of annotated data but also achieves excellent detection performance as the number of annotated data increases.
\section{Discussion}
\subsection{Contributions to clinical fields}
\textcolor{black}
{
Different methods are available for the evaluation of gastritis: double-contrast upper gastrointestinal barium X-ray radiography, upper gastrointestinal endoscopy, and serum anti-$H.\,pylori$ IgG titer~\cite{togo2019detection}.
Among such methods, the screening of gastric X-ray images is still the most simple and widely used method, and thus, it is suitable for mass screening in East Asia.
However, the number of clinicians who can perform radiological diagnosis is decreasing owing to diverse inspection approaches~\cite{sugano2015screening}.
Therefore, the introduction of DCNN-based CAD systems is expected in gastritis detection.
}
\par
\textcolor{black}{
As shown in our previous studies, gastritis can be detected from gastric X-ray images with high reliability by DCNN-based CAD systems~\cite{togo2019detection}.
Similar to most medical image analysis studies, our previous methods are based on supervised learning~\cite{kanai2019gastritis}. 
The performance of our previous methods depends on a large number of manually annotated gastric X-ray images.
Generating annotations of complex gastric X-ray images typically requires expert knowledge.
Thus, it is expensive and time-consuming~\cite{zhou2021models}.
The proposed self-supervised learning method addressed this challenging problem and achieved a high detection performance with only a few annotated gastric X-ray images.
The method can reduce the labor required for annotating gastric X-ray images.
Although our research focuses on gastric X-ray images, we expect that the findings and methodology of this study can be applied to other types of medical images such as laparoscopic, endoscopy, and computed tomography images.
Furthermore, because the self-supervised learning process is task-agnostic, we can also apply the trained encoder to other tasks, such as few-shot learning, semantic segmentation, and anomaly detection.
}
\par
\textcolor{black}{
Our research has several limitations.
The experimental data used in this study were obtained from a single medical facility, which could affect the robustness of our method.
Several studies have shown that additional data enable increasing the representation learning performance in self-supervised learning~\cite{zhou2021models}.
In the future, we will explore the effect of data from multiple medical facilities on self-supervised learning.
Our previous study~\cite{li2020soft, li2022compressed, li2023sharing} about dataset distillation can improve the effectiveness and security of medical data sharing among different medical facilities, which fits well with self-supervised learning.
}
\subsection{Contributions to technical fields}
Early self-supervised learning methods usually learn discriminative representations based on pretext tasks.
For example, the study~\cite{gidaris2018unsupervised} predicts the rotation degrees of images, and discriminative representations were learned by playing a jigsaw game on images~\cite{noroozi2016unsupervised}.
Because these pretext tasks are monotonous and not complex, sufficient discriminative representations cannot be learned from images~\cite{liu2021self}.
PIRL~\cite{misra2020self} proposes combining the jigsaw or rotation pretext task with learned invariant representations, outperforming supervised learning in certain tasks.
\par
Recently, several self-supervised learning methods have achieved excellent performance on the large-scale natural image dataset ImageNet~\cite{deng2009imagenet}.
Specifically, SimSiam~\cite{chen2021exploring} and BYOL~\cite{grill2020bootstrap} perform self-supervised learning by directly reducing the distance between the representations of two views from the Siamese networks.
These methods are efficient for processing gastric X-ray images with high-resolution and complex semantic information.
Therefore, we propose a novel self-supervised learning method based on a \textcolor{black}{teacher-student} architecture (a variant of the Siamese networks) for gastritis detection using gastric X-ray images.
We introduce cross-view and cross-model losses, which can perform explicit self-supervised learning and learn discriminative representations from gastric X-ray images~\cite{li2021cross}. 
To the best of our knowledge, this study is the first to prove the effectiveness of self-supervised learning in the field of gastritis detection using gastric X-ray images.
Experimental results show that the PM can achieve a high patient-level detection performance in gastritis detection with only a few annotations.
\section{Conclusion}
We propose a novel self-supervised learning method based on a \textcolor{black}{teacher-student} architecture for gastritis detection using gastric X-ray images.
\textcolor{black}
{
We present cross-view and cross-model losses that enable explicit self-supervised learning and learn discriminative representations from gastric X-ray images.
}
The experimental results showed that the proposed method achieved a high patient-level gastritis detection performance with only a few annotations.
\section*{Ethical approval}
No ethics approval is required.
\section*{Declaration of competing interest}
None declared.
\section*{Acknowledgments}
We express our thanks to Nobutake Yamamichi of the Graduate School of Medicine, The University of Tokyo, and Katsuhiro Mabe of the Junpukai Health Maintenance Center.
\section*{Fundings}
This study was supported in part by AMED Grant Number JP21zf0127004, the Hokkaido University-Hitachi Collaborative Education and Research Support Program, and the MEXT Doctoral program for Data-Related InnoVation Expert Hokkaido University (D-DRIVE-HU) program. 
This study was conducted on the Data Science Computing System of Education and Research Center for Mathematical and Data Science, Hokkaido University. 
\bibliographystyle{spmpsci}
\bibliography{refs}

\end{document}